# ArTST: Arabic Text and Speech Transformer

**Hawau Olamide Toyin,**[*] **Amirbek Djanibekov,**[*] **Ajinkya Kulkarni, Hanan Aldarmaki**
Mohamed bin Zayed University of Artificial Intelligence
Abu Dhabi, UAE
{hawau.toyin;amirbek.djanibekov;ajinkya.kulkarni;hanan.aldarmaki}@mbzuai.ac.ae

## Abstract

We present ArTST, a pre-trained Arabic text and speech transformer for supporting open-source speech technologies for the Arabic language. The model architecture follows the unified-modal framework, SpeechT5, that was recently released for English, and is focused on Modern Standard Arabic (MSA), with plans to extend the model for dialectal and code-switched Arabic in future editions. We pre-trained the model from scratch on MSA speech and text data, and fine-tuned it for the following tasks: Automatic Speech Recognition (ASR), Text-To-Speech synthesis (TTS), and spoken dialect identification. In our experiments comparing ArTST with SpeechT5, as well as with previously reported results in these tasks, ArTST performs on a par with or exceeding the current state-of-the-art in all three tasks. Moreover, we find that our pre-training is conducive for generalization, which is particularly evident in the low-resource TTS task. The pre-trained model as well as the fine-tuned ASR and TTS models are released for research use.

## 1 Introduction

Large pre-trained transformer models are currently at the forefront of speech and text technologies, with applications in various text and speech recognition and generation tasks (Devlin et al., 2019; Raffel et al., 2020; Hsu et al., 2021; Baevski et al., 2020). These models share several aspects: (1) they are based on the transformer architecture (Vaswani et al., 2017), which enables efficient training of larger models and incorporating wider contexts, (2) they are scaled in terms of model size, which has been shown to correlate with performance (Alabdulmohsin et al., 2022; Hestness et al., 2017), and (3) they generally use a self-supervised training objectives, such as next token prediction (Brown et al., 2020), masked prediction (Devlin et al., 2019; Hsu et al., 2021), and contrastive loss (Baevski et al., 2020), which enable the utilization of large unlabeled datasets for multiple potential downstream tasks. Pre-trained self-supervised models like Wav2Vec2.0 (Baevski et al., 2020), and its multi-lingual variant (Babu et al., 2022), have mostly replaced traditional acoustic features like MFCCs and filter banks in the speech domain. These pre-trained models implicitly learn robust and generalizable acoustic representations that consistently improve performance in various supervised downstream tasks with acoustic inputs like Automatic Speech Recognition (ASR). This is achieved by simply adding a prediction layer and fine-tuning the model using a suitable loss function, such as CTC loss (Graves, 2012).

This *pre-train-then-finetune* framework is flexible for a variety of applications, but most pre-trained models are uni-modal and therefore are limited to tasks that share the same input modality. For instance, acoustic models like Wav2Vec2.0 are not typically used in text-to-speech synthesis applications, where the input is text, and the output is typically in the form of mel spectrograms. For this reason, self-supervised pre-training has not been as widely adopted in speech synthesis research. One exception to this trend is the SpeechT5 model (Ao et al., 2022), which accepts both text and speech as input and output using modal-specific networks in addition to the core encoder-decoder network. The model is first pre-trained using self-supervised objectives in both text and speech modalities, and then fine-tuned on a variety of supervised tasks, including speech transcription, speech synthesis, and speech classification. SpeechT5 has been trained only on English using more than 900 hours of speech and 400 million sentences of text data. While the model can technically be fine-tuned for other languages, our preliminary evaluations of Arabic fine-tuning show poor performance; the pre-training seems to have biased the model severely for recognizing and generating English speech.

---

[*] These authors contributed equally to this work.

In this paper, we introduce **Ar**abic **T**ext and **S**peech **T**ransformer, **ArTST**[1], a project aiming to push the boundaries for Arabic open-source speech technology by providing various pre-trained speech and text transformers. The Arabic language exhibits significant dialectal variation and code-switching, which introduce a layer of complexity for speech recognition and generation tasks. We believe this can be best addressed via methodical and focused development of self-supervised models that target this linguistic landscape rather than multi-lingual models that may compromise mono-lingual performance for multi-lingual coverage. The first release, as described in this paper, is a direct adaptation of the SpeechT5 model, but pre-trained from scratch using Modern Standard Arabic data and evaluated on various downstream tasks. Future versions will include dialectal Arabic, as well as code-switched speech and text, by exploring the best architectural modifications for improving coverage without sacrificing performance for individual variants.

We demonstrate the performance of ArTST in the following tasks: Automatic Speech Recognition (ASR), Text-To-Speech synthesis (TTS), and spoken Dialect Identification (DID). The fine-tuned models on each task achieved performance on a par with or exceeding previously reported results on our test sets, establishing a new state-of-the-art for open-source models. For ASR, the model additionally outperforms the large pre-trained ASR models, `Whisper` (Radford et al., 2023), and `MMS` (Pratap et al., 2023), which further demonstrates the advantage of focusing only on Arabic. Moreover, we report some interesting findings in TTS fine-tuning, as the model learns to synthesize speech without explicit text diacritization in a way that generalizes to unseen domains, which we believe is a result of the unsupervised pre-training on large Arabic speech data. Our main contributions are:

1. Releasing a pre-trained cross-modal transformer model capable of handling diverse speech and text tasks, in addition to fine-tuned ASR and TTS models for MSA[2].

2. Demonstrating state-of-the-art performance in ASR, TTS, and DID, using standard open-domain datasets for MSA.

3. Demonstrating unique generalization capabilities, such as speech synthesis without explicit diacritization.

## 2 Related Works

To the best of our knowledge, there is no model pre-trained on Arabic that can perform multiple downstream speech-related tasks with different input modalities. In the text domain, AraT5 (Elmadany et al., 2022) was implemented as an Arabic version of the Text-To-Text Transfer Transformer (T5) model (Raffel et al., 2020), which uses transfer learning with a unified Transformer framework for several downstream text generation tasks. In the speech domain, multi-lingual acoustic models, such as XLSR-R (Babu et al., 2022), Whisper (Radford et al., 2023), or MMS (Pratap et al., 2023), include Arabic as one of many languages in supervised or self-supervised pre-training, but they can only handle speech as input modality, and text as output modality. ArTST is directly inspired from the SpeechT5 model (Ao et al., 2022), which is a pre-trained encoder-decoder transformer with additional modal-specific networks to handle both text and speech modalities in the input and output. The model was shown to be versatile as it can achieve superior performance when fine-tuned for ASR, TTS, and other speech related tasks. However, the model was pre-trained only on English data, and as a result, the internal representations seem to be heavily biased towards English speech. By fine-tuning the model for Arabic ASR and TTS, our experiments indicate that it may be difficult to overcome this bias without multi-lingual pre-training.

Several studies attempted to measure the effect multi-lingual pre-training in acoustic models, with mixed results (Yadav and Sitaram, 2022). Heigold et al. (2013) compared models pre-trained on English only with models trained on multi-lingual data using conventional HMM-DNN models, and showed empirically that multilingual pre-training is better than fine-tuning an English model on a different target language. Huang et al. (2013) further shows that multilingual pre-trained features can generalize to unseen languages. Tong et al. (2017) shows that multi-lingual ASR training is worse than monolingual training in the target language, but multilingual pre-training followed by target language fine-tuning is better than monolingual training. Language similarity likely plays a role in generalization: Ram and Aldarmaki (2022) showed that acoustic word embeddings obtained using Wav2Vec 2.0 features that are pre-trained on English generalize to languages like French and German, but don't generalize as well for Arabic. Furthermore, several studies show that multilingual models generalize better using language vectors or language adapters (Kannan et al., 2019; Toshniwal et al., 2018; Shetty and NJ, 2020; Radford et al., 2023; Pratap et al., 2023), which indicates that some language-specificity in the model is preferable to crude multi-lingual training. Some empirical evidence also suggests that performance of some high-resource languages can potentially degrade in multi-lingual settings compared to monolingual pre-training (Watanabe et al., 2017).

---

[1]Pronounced 'artist'.
[2]https://github.com/mbzuai-nlp/ArTST

The above mentioned studies all focus on acoustic models where speech is the input rather than the output. Text-to-speech synthesis models, on the other hand, are generally more fragile and highly depend on the quality and size of training data. Generally speaking, TTS models require consistent and clean recordings in order to synthesize natural and intelligible speech (Kulkarni et al., 2023). Multi-lingual TTS synthesis is an emerging topic of research, but these attempts are rare compared to multi-lingual ASR and cover only a small subset of languages due to shortage of resources suitable for speech synthesis (Li et al., 2021; Cho et al., 2022).

## 3 ArTST

ArTST is a text and speech transformer optimized for the Arabic language. Based on observations from previous studies on multilingual and monolingual ASR, TTS, and self-supervised pre-training, we believe that training a model from scratch with the Arabic language in mind would improve the quality of the resulting models. Our strategy is to start with a monolingual setting, and explore the optimal settings for Modern Standard Arabic (MSA) speech processing. In future iterations of the model, we will explore how best to expand it to handle various dialects as well as other languages that are often mixed with Arabic (i.e. English and French). We believe that an incremental approach of this kind is more likely to lead to optimal performance. Here, we describe the first stage of this project, which focuses only on MSA. ArTST is adapted from the transformer-based SpeechT5 model, which we briefly describe in this section. For more details, please refer to Ao et al. (2022).

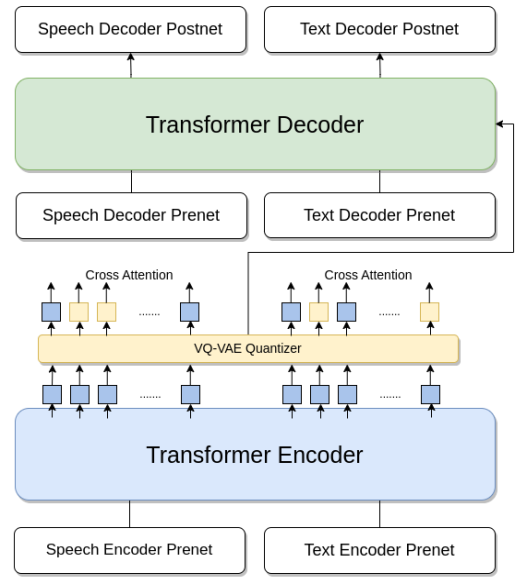

Figure 1: Model architecture.

### 3.1 Model Architecture

Figure 1 shows the overall architecture of the model. It consists of a main encoder-decoder transformer network, similar to the architecture employed in T5 (Raffel et al., 2020). This network is shared for both speech and text modalities. To account for the differences in pre- and post-processing, additional modal-specific pre- and post-nets are used to handle the text and speech features.

### 3.2 Pre-training

The model is pre-trained using various self-supervised objectives to account for both speech and text modalities in the input and output:

**Speech bidirectional masked prediction**: Following the framework of HuBERT (Hsu et al., 2021), discrete frame-level targets are employed for masked prediction, where random spans of 10 steps from the output of the speech encoder pre-net are masked across each utterance, and the model is trained to predict the correct discrete labels via cross-entropy. The discrete labels are obtained from a pre-trained HuBERT model (Hsu et al., 2021), where the hidden representations are clustered into 500 classes using the k-means algorithm. This training objectives can be a stepping stone towards learning speech to text transformation as the model is trained to map continuous speech features into discrete units. This objective updates the speech encoder pre-net as well as the main encoder.

**Speech de-noising auto-encoder**: This objective trains the speech decoder pre-net, decoder, and speech decoder post-net to reconstruct speech features in the form of 80-dimensional log mel filterbanks from the randomly masked utterances as described above.

**Text de-noising auto-encoder**: Using unlabeled text, the text encoder pre-net, encoder-decoder network, and text decoder pre- and post-nets, are all optimized using a denoising reconstruction loss.

**Cross-modal loss**: Vector-quantized embeddings are used to implicitly align speech and text representations through a shared code-book. During training, 10% of the contextual embeddings are replaced with the corresponding quantized embeddings, and the cross-attention in the main encoder-decoder transformer is calculated based on this mixed representation. A diversity loss is used to encourage sharing more codes between the text and speech inputs.

In ArTST, each of the encoder and decoder components are similar in size and configuration to SpeechT5 (Ao et al., 2022). Speech pre/post-nets and text pre/post-nets all have the same structure as in the SpeechT5 model, with the only difference being in the text tokenizer which we initialize using the characters in our training sets. We employed the official HuBERT model[3] to generate the discrete labels for the bidirectional masked prediction objective since a pre-trained Arabic HuBERT model was not available for our perusal. In future work, we will explore the potential of improving this component using a model pre-trained on Arabic speech.

### 3.3 Fine-Tuning

Task-specific fine-tuning is carried out by employing the encoder-decoder backbone in addition to the relevant pre- and post-nets. For example, for ASR, the speech encoder pre-net, and text decoder pre- and post-nets are used to handle speech input and text output. All relevant model parameters are updated during fine-tuning.

---

[3]https://github.com/facebookresearch/fairseq/blob/main/examples/hubert

## 4 Training & Fine-Tuning Settings

### 4.1 Dataset

For training our MSA ArTST model, we utilize the Multi-Genre Broadcast (MGB2) dataset (Ali et al., 2016), which is collected from Aljazeera TV recordings of Arabic speech, mostly in MSA. This dataset is often used for benchmarking ASR models for MSA, which enables fair comparison with previous research. The original dataset contains 1.4K unique speakers with ∼1.2K hours of transcribed speech data. We excluded overlapping speech utterances from the set, which are tagged in the corpus. Furthermore, to avoid high amount of padding and maintain a balance between computational efficiency and effectiveness, we excluded speech samples that exceeded a duration of 40 seconds. The resulting dataset consists of roughly 1K hours of speech. We also randomly extracted a 200 hr subset of MGB2 for the purpose of performing preliminary experiments to evaluate SpeechT5 fine-tuning on ASR. Moreover, we extracted a random subset from the QASR corpus (Mubarak et al., 2021), a multi-dialectal broadcast speech corpus from Aljazeera that includes MSA speech as well as dialectal Arabic of different varieties. As we are focusing mainly on MSA in this work, we do not utilize this dataset for pre-training, but instead utilize it to test the generalization potential of the model. For TTS fine-tuning, we utilize open-source Arabic datasets curated for speech synthesis, namely: The Arabic Speech Corpus (ASC) (Halabi et al., 2016) and Classical Arabic Text-to-Speech Corpus (ClArTTS) (Kulkarni et al., 2023)[4]. We also utilize these two datasets for evaluating the ASR models. For all datasets, we use the predefined test/dev splits if applicable. We summarize all dataset statistics in Table 1.

### 4.2 Text & Speech Pre-Processing

All punctuation marks were removed with the exception of @ and %. Additionally, all diacritics were removed, and Indo-Arabic numerals were replaced with Arabic numerals to ensure uniformity. The vocabulary is comprised of individual Arabic alphabets, numerals, and select English characters from the training dataset, in addition to some special characters like @ and %. For speech data, we standardized the sampling rate to be 16 kHz across all collected datasets.

---

[4]www.clartts.com

| | Split | # of Hours | # of Words |
|---|---|---|---|
| MGB2 | MGB2-1K (train) | 1005.39 | 6.96M |
| | MGB2-200 (train) | 201.32 | 1.39M |
| | test | 9.57 | 64.38K |
| QASR | QASR-267 (train) | 267.91 | 2.00M |
| | test | 9.57 | 64.38K |
| ASC | train | 3.81 | 20.58K |
| | test | 0.28 | 1.40K |
| ClArTTS | train | 11.16 | 76.27K |
| | test | 0.24 | 1.69K |

Table 1: Datasets used in our experiments.

| | Train set / Test set | Enc | WER↓ | CER↓ |
|---|---|---|---|---|
| SpeechT5 | ASC / ASC | Ar | 78.07% | 23.54% |
| | | Bw | 76.92% | 22.02% |
| ArTST | ASC / ASC | Ar | 45.8% | 9.88% |
| SpeechT5 | ClArTTS / ClArTTS | Ar | 32.31% | 6.88% |
| | | Bw | 24.32% | 5.12% |
| ArTST | ClArTTS / ClArTTS | Ar | 12.51% | 3.60% |
| SpeechT5 | MGB2-200 / MGB2 | Ar | 69.74% | 26.47% |
| | | Bw | 45.09% | 17.55% |
| ArTST | MGB2-200 / MGB2 | Ar | 16.56% | 7.68% |
| SpeechT5 | QASR-267 / MGB2 | Ar | 72.70% | 26.27% |
| | | Bw | 53.19% | 19.01% |
| ArTST | QASR-267 / MGB2 | Ar | 17.27% | 9.99% |

Table 2: Fine-tuned ASR resutls using English SpeechT5 vs. ArTST in terms of Word Error Rate (WER) and Character Error Rate (CER). Character Encoding (Enc): Arabic (Ar), BuckWalter (Bw).

### 4.3 ArTST Pre-training

We pre-trained ArTST using the MGB2-1K subset. Since the pre-training is unsupervised, aligned text and speech data are not required at this stage. For text pre-training, we employed the cleaned transcriptions from the MGB2 dataset as unlabeled data. We pre-trained ArTST using Adam optimizer with a learning rate of $2 \times 10^{-4}$, spanning 200K updates, and a warm-up phase of 64K updates. The maximum speech token length was set at 250K (equivalent to 15.625 seconds), and the text tokens were capped at 600 characters. The pre-training was run on four A100 GPUs for 14 days.

## 5 Results & Evaluation

### 5.1 SpeechT5 Finetuning vs. ArTST

We conducted preliminary assessments of the SpeechT5 model from Ao et al. (2022), which was pre-trained and fine-tuned on English, to assess the ability of cross-lingual transfer by directly fine-tuning the model for Arabic ASR using various Arabic speech datasets. We experimented with both the original Arabic script as input, as well as Buckwalter transliteration (Habash et al., 2007) instead of Arabic script to account for the fact that the model was pre-trained only on English characters.

For Arabic script, we augmented the original character tokenizer to incorporate symbols that correspond to Arabic letters and special symbols contained in the fine-tuning set. The original tokenizer contained approximately 80 symbols; after incorporating the Arabic letters and special symbols, the extended tokenizer vocabulary increased to 130 symbols. Furthermore, we modified the input embeddings structure to align with the dimensionality of the updated tokenizer. The embedding layer retains the weights from the earlier-trained SpeechT5 model for its initial 80 components. Meanwhile, additional elements were initialized randomly. Similarly, for Buckwalter transcriptions, we modified the tokenizer accordingly. Since the transliteration scheme contains mostly English alphabets in addition to some special ASCII characters, the extended vocabulary in this setting was increased to 90 characters. We start with the pre-trained English ASR from SpeechT5[5] and fine-tune it on the specified datasets until training and validation loss diverge.

Table 2 shows the results in terms of Word Error Rate (WER) and Character Error Rate (CER) in all different settings. The ArTST ASR model was fine-tuned using our pre-trained ArTST using the same tokenizer as the pre-trained model, which contains Arabic script.

**Effect of Input Encoding**

We see from these experiments that SpeechT5 fine-tuning is improved using Buckwalter rather than Arabic script. Since the transcription scheme mostly results in mapping Arabic letters to similar-sounding English letters, the learning objective does not diverge greatly from the original English model, which results in improved performance compared to using Arabic script. In our analysis, approximately 85% of Arabic characters were replaced with corresponding English characters, facilitating the continuation of fine-tuning for SpeechT5's ASR, even with limited data.

---
[5]huggingface.co/microsoft/speecht5_asr

| Model | WER↓ | CER↓ |
|---|---|---|
| From (Hussein et al., 2022): | | |
| HMM-DNN | 15.80% | — |
| E2E, CTC + LM | 16.90% | — |
| E2E, Attention + LM | 13.40% | — |
| E2E, CTC , Attention + LM | **12.50%** | — |
| ArTST | 13.42% | 6.43% |
| ArTST + LM | **12.78%** | **6.33%** |

Table 3: Comparing ArTST performance against models reported in (Hussein et al., 2022), which include best performing model previously reports on MGB2.

**Effect of Pre-Training**

We also observe large reductions in error rate using the same datasets for fine-tuning ArTST. The difference in performance is evident in all cases, but it's particularly large for the ASC and MGB2-200 subsets. SpeechT5 fine-tuned with Buckwalter transcriptions on the ClArTTS corpus results in relatively good performance of 24% WER compared to 12.78% WER for ArTST. For the other two datasets, the difference is roughly 30% absolute WER in favor of ArTST. This could potentially be resulting from two factors: ClArTTS is a consistent and clean dataset that was curated for TTS, compared to MGB2 which is extracted from TV shows. ASC is also curated for TTS, and therefore consists of clean and consistent recordings, but dataset size could have played a role in the high WER for ASC, which is much smaller than the ClArTTS dataset (∼3.8 hrs compared to ∼11.16 hrs). While MGB2 contains orders of magnitude more data than ClArTTS, the error rates are higher than ClArTTS for all models, including ArTST, which is further evidence that dataset quality is most likely playing a role in these results.

Finally, we also used a subset of QASR for fine-tuning ASR models as a counterpoint for the MGB2 datasets because the latter was used in pre-training and could have biased the results in favor of ArTST. However, even in this set, we clearly see that ArTST performs much better than the fine-tuned SpeechT5, with error rates on a par with the ones observed for MGB2.

## 5.2 Benchmarking ArTST for MSA

We fine-tuned ArTST on our MGB2-1K dataset, and compared the performance against comparable models trained and tested on MGB2. Since 2017, the lowest WER on MGB2 test set was reported in Smit et al. (2017) as 13.2%. Recently, Hussein et al. (2022) explored the potential of an end-to-end transformer model compared to conventional ASR systems, and achieved state-of-the-art performance in the MGB2 test set. The model was trained on the MGB2 dataset, so it's comparable to our model in that regard. Furthermore, they utilize a language model for rescoring using the MGB2 transcriptions as well as the additional 130M words of text data provided in the MGB2 challenge. Our model consists of the speech pre-net, encoder, and text pre/post-nets fine-tuned with CTC loss. We also experiment with LM shallow fusion using a transformer-based auto-regressive character language model trained on the same sets. We used the default LM setting from the Fairseq library[6], and we trained the model for 300K updates using the Adam optimizer, with 4K warm-up steps, a learning rate of 0.0005, and 0.1 dropout rate.

The results are shown in Table 3. Our model without LM fusion achieves 13.42% WER, which is on a par with the transformer-based end-to-end model with attention and LM rescoring reported in Hussein et al. (2022). Furthermore, ArTST outperforms the architecture most similar to it (E2E, CTC + LM) by more than 3% absolute WER, without incorporating a language model for inference. The error rates are further reduced to 12.78% by incorporating LM fusion, which is comparable to the best model reported in Hussein et al. (2022); the latter incorporates both Attention and CTC, as well as LM rescoring with beam size of 20.

## 5.3 Comparing ArTST With Multilingual Models

Recently, a few large multi-lingual pre-trained models have been released for ASR in multiple languages, such as Whisper (Radford et al., 2023), and MMS (Pratap et al., 2023). Both models include Arabic as one of many languages included in their supervised pre-training. Training data, model architectures, training objectives, and model sizes vary considerably between these models, so they are not directly comparable, However, the fact that these models are widely circulated and used necessitates some kind of performance comparison with our model.

Table 4 shows the WER/CER of these models in Arabic ASR using our test sets. We also report the number of parameters for each model.

---
[6]github.com/facebookresearch/fairseq

| Test Set | ArTST | | Whisper$_{medium}$ | | Whisper$_{large}$ | | MMS$_{medium}$ | | MMS$_{large}$ | |
|---|---|---|---|---|---|---|---|---|---|---|
| | WER | CER | WER | CER | WER | CER | WER | CER | WER | CER |
| **ASC** | 45.70% | 9.73% | 48.46% | 10.74% | 47.73% | 10.83% | 54.05% | 11.71% | 57.37% | 11.13% |
| **ClArTTS** | 13.52% | 3.90% | 20.49% | 6.24% | 19.25% | 6.23% | 36.18% | 9.17% | 31.13% | 6.58% |
| **MGB2** | 13.42% | 6.43% | 28.69% | 11.72% | 26.71% | 10.78% | 45.58% | 14.86% | 40.33% | 13.06% |
| **QASR(1hr)** | 26.08% | 16.65% | 36.54% | 17.45% | 32.32% | 15.56% | 52.79 % | 20.86% | 47.81 % | 18.80% |
| # params | 155 M | | 769 M | | 1550 M | | 300 M | | 965 M | |

Table 4: ArTST compared with large multi-lingual models: Whisper & MMS on our test sets. ArTST was fine-tuned for ASR using MGB2-1k train set. Results are shown without LM fusion.

While Whisper performs relatively well compared to MMS, ArTST outperforms both models, including the large variant of each model, in all test sets, while having a smaller number of parameters. For instance, without LM fusion, ArTST achieved 13.5% WER on MGB2 test set, while the large variants of Whisper and MMS achieved 26.7% and 40.3% WER, respectively.

## 5.4 Qualitative Analysis of ASR Output

In Table 5, we show some examples of ASR outputs from ArTST compared with the reference transcriptions. These examples show the drawback of the raw WER/CER metrics as they don't account for potential variations in spelling. In particular, we observed several cases where English words are transliterated or misspelled. Furthermore, numeric expressions, like 80%, are in some cases written in numeric format, and others spelled out in words. Furthermore, the large error rates reported for ASC are in a large part caused by intentional misspelling in the reference ASC transcriptions, which are intended to facilitate learning of TTS synthesis in a low-resource setting. In the shown examples, ArTST output is in fact the correct spelling. We also show a couple of examples of ArTST, which is fine-tuned on MGB2, generalizing to dialectal Arabic utterances from QASR.

## 5.5 ArTST for TTS Synthesis

We experimented with TTS fine-tuning, comparing ArTST pre-trained model with SpeechT5 TTS[7] as a starting point. We fine-tuned each model using the ClArTTS and the ASC datasets, which are two open-source datasets curated for Arabic TTS. For the SpeechT5 model, we used Buckwalter transcriptions for the text, as our experiments in ASR demonstrated it to be more suitable for this model. For both models, we fine-tuned the TTS model without using input diacritics, so no automatic diacritizer is needed for inference. This feature diverges from previous works in Arabic TTS, where efforts are taken to include diacritization in the input text. However, since this would necessitate the use of text-based diacritizers for inference, and as shown in Aldarmaki and Ghannam (2023), text-based diacritizers have high error rates when applied to the speech domain. We opted to train undiacritized TTS instead, and let the model implicitly learn the correct pronunciation.

| Examples from MGB2 Test |
|---|
| أنه إذا أنت تطرح الأبرتهايد يجب أن تطرح حالة من المساواة للجميع |
| أنه إذا أنت تطرح الـ apartheid يجب أن تطرح حالة من المساواة للجميع |
| وتشرفهم وتكرمهم بل في الثمانين بالمائة الذين لم ينجحوا لا هم معدون لشيء |
| وتشرفهم وتكرمهم بل في 80% الذين لم ينجحوا لا هم معدون لشيء |
| **Examples from ASC Test** |
| وبالتالي تساعد على لوقاية من لإمساك |
| وبالتالي تساعد على الوقاية من الإنساك |
| وذلك على خلاف نظرائه لسابقين |
| وذلك على خلاف نظرائه السابق |
| **Examples from QASR Test** |
| أنه كأنه والله الجندي السوري السني الذي يقاتل هو فقط يقاتل خوفا |
| أنه كأنه والله الجندي السوري السني الذي يقاتل هو فقط يقاتل خوفا أنا لا |
| وا لأولاد عسكوا و يقعدوا هناك في الجنينة عشان يعملوا بعض مظاهر تتعلق بالثورة |
| والولاد عسكوا يقعدوا هناك في الجنينة عشان يعملوا بعض مظاهر تتعلق بالثورة |

Table 5: Sample ArTST ASR transcriptions (bottom) vs. reference transcriptions (top). Highlighting differences or errors. Correct words not present in ArTST or reference. Correct words in ArTST but not present in reference.

The fine-tuning was carried out using the text encoder pre-net, encoder/decoder backbone, and speech decoder pre/post-nets. All model parameters were updated during fine-tuning. We used the pre-trained HiFi-GAN vocoder[8] to convert the output of each model to waveform.

---

[7] huggingface.co/microsoft/speecht5_tts

[8] huggingface.co/microsoft/speecht5_hifigan

|  | Fine-tuning Data | MOS ↑ |
|---|---|---|
| Ground Truth | ASC | 4.31 |
|  | ClArTTS | 4.64 |
| English SpeechT5 | ASC | 1.57 |
|  | ClArTTS | 1.88 |
| ArTST | ASC | 2.93 |
|  | ClArTTS | 4.11 |
| ArTST* | ASC | 3.44 |
|  | ClArTTS | 4.31 |

Table 6: Subjective listening tests in terms of Mean Openion Score (MOS) for models fine-tuned using English SpeechT5 vs. ArTST, vs. ArTST* (variant of TTS model pre-trained on MGB-2 data).

### TTS Pre-Training

Since both ASC and ClArTTS are relatively small datasets, we also experimented with TTS pre-training using ASR data from MGB2. Generally speaking, ASR data are not suitable for TTS training due to the high variability is speaking style and presence of noise. On the other hand, ASR data are available in abundance, and can potentially help improve the model's generalization potential. We start by fine-tuning the TTS model on MGB2-1K train set, and then fine-tune it again on the TTS train sets. We refer to this variant as ArTST*.

### TTS Evaluation

We conducted subjective evaluation through listening tests to assess the naturalness and intelligibility of the synthesized speech from differnet models in a single score from 1 to 5 (higher is better). We selected random utterances from each test set, and synthesized speech based on the corresponding text transcription using the variants speechT5, ArTST and ArTST*. Fifteen native Arabic speakers participated in the evaluation. The Mean Openion Score (MOS) for each model is shown in Table 6. The audio samples used in the evaluation are available here [9]. As seen from the table, and through the provided samples, using the pre-trained SpeechT5 model as a basis for fine-tuning leads to very poor speech synthesis. On the other hand, using the pre-trained ArTST as a basis for fine-tuning results in high-quality synthesis. Furthermore, pre-training the TTS model using MGB2 ASR data further improves the quality of the transcriptions. Moreover, we observed through listening tests that the model generalizes to unseen sentences from MSA, where we synthesized speech from transcriptions obtained from QASR[10]. In particular, the model learns to produce the correct pronunciation in spite of not being provided with any diacritics.

[9] https://artstts.wixsite.com/artsttts

| Model | Dev | Test |
|---|---|---|
| E2E (softmax) (Shon et al., 2020) | 83.00% | 82.00% |
| HuBERT-17 (Sullivan et al., 2023) | 92.23% | 92.12% |
| XLS-R-300M-17 (Sullivan et al., 2023) | 90.77% | 90.20% |
| ArTST | **95.08%** | **94.18%** |
| *MGB-5 Challenge (Ali et al., 2019) Top 2 Systems:* | | |
| UKent | 93.50% | 93.10% |
| DKU [Single best system] | 94.70% | 93.80% |
| DKU [Fusion of 4 systems] | 97.40% | 94.90% |

Table 7: Accuracy results for dialect identification on the ADI17 set.

### 5.6 Dialect Identification

To fine-tune ArTST for speech classification, we recast the multi-class classification task as a speech to text generation task. The decoder is then trained to predict the dialect class at the first time step (which is equivalent to a regular softmax classifier). We fine-tuned all parameters using the Arabic Dialect Identification for 17 countries (ADI17) dataset (Shon et al., 2020). We compared our model to previously reported results in Table 7. As seen from these results, ArTST outperforms previous models, including the best single system submitted to the MGB-5 challenge (Ali et al., 2016), and is not far behind the top model which fuses 4 different system; it is worth noting that the latter also incorporates data augmentation to further improve performance, which we do not explore in this work.

## 6 Conclusions & Future Work

We demonstrated the potential of ArTST in speech recognition, synthesis, and classification, where we achieved results on a par with or outperforming previously reported results with relatively straightforward fine-tuning. What we have demonstrated in this paper is only a subset of potential applications of this framework. As the model can handle both text and speech modalities, it can potentially be applied for text-to-text and speech-to-speech applications, in addition to text classification and generation tasks. We will explore these avenues of application in future work. In this initial work, we focused on MSA as the main variant of Arabic for pre-training. We explored the potential of the model to generalize to dialectal Arabic using small test sets that include dialectal Arabic, as well as the dialect identification task. Future edition will focus on expanding the coverage of the pre-trained model to include various dialects, and potentially code-switched speech, without sacrificing performance on MSA. As demonstrated in this paper, our model outperforms larger multi-lingual models like Whisper and MMS, which we believe is a result of focusing on the Arabic language as a basis of our model from its inception. While multi-linguality may be desirable for some applications, and could be beneficial for low-resource languages, monolingual models have a greater potential for high-resource languages, and the Arabic language currently boasts large volumes of open-source datasets that can be utilized to develop high-quality models across various tasks.

[10]Samples are available in the same website.

## 7 Limitations

As this is a large on-going project comprising several tasks and potential variations in pre-training, there are several limitations that can be acknowledged here. First, the model's pre-training consists of mainly MSA speech from a single dataset (MGB2). While this dataset is large and comparable to the pre-training conditions in SpeechT5, there are other datasets that could be incorporated to potentially improve performance. Furthermore, we did not focus on dialectal Arabic in this edition, and only alluded to potential generalization to dialects through some experiments on ASR and dialect identification. Given small amount of code-switching in the MGB2 set, the model does have limited code-switching recognition, but it can be improved by intentionally using code-switching dataset for pre-training and fine-tuning. One more limitation is the use of pre-trained HuBERT for generating intermediate discrete labels in the pre-training stage. While our model demonstrably achieves excellent results in all tested tasks in spite of that, we did not explore the possibility of optimizing HuBERT for Arabic, mainly due to the additional computational load for training another large model. Finally, we did not probe the internal representations of the model to explore potential architectural improvements. Further analysis of these representations, and a thorough analysis of the dialect identification model could shed light on the properties of these representations.